\documentclass{article}

\usepackage{microtype}
\usepackage{graphicx}
\usepackage{subfigure}
\usepackage{booktabs} 
\usepackage{pifont}
\usepackage{hyperref}

\usepackage[accepted]{icml2024}

\usepackage{amsmath}
\usepackage{amssymb}
\usepackage{mathtools}
\usepackage{amsthm}
\usepackage{newfloat}
\usepackage{listings}
\usepackage{algorithm}
\usepackage{algorithmic}
\usepackage{adjustbox}
\usepackage{multirow}
\usepackage{booktabs}
\usepackage{tabularx}
\usepackage{color}
\usepackage{bm}
\usepackage{amsmath}
\usepackage{tikz}
\usepackage{xcolor}
\usepackage{tcolorbox}
\usepackage{wrapfig}

\usepackage[capitalize,noabbrev]{cleveref}

\theoremstyle{plain}

\theoremstyle{definition}

\theoremstyle{remark}

\usepackage[textsize=tiny]{todonotes}

\icmltitlerunning{Auto-Encoding Morph-Tokens for Multimodal LLM}

\begin{document}

\twocolumn[
\icmltitle{Auto-Encoding Morph-Tokens for Multimodal LLM}



\icmlsetsymbol{equal}{*}

\begin{icmlauthorlist}
\icmlauthor{Kaihang Pan}{zju}
\icmlauthor{Siliang Tang}{zju}
\icmlauthor{Juncheng Li}{zju,nus}
\icmlauthor{Zhaoyu Fan}{zju}
\icmlauthor{Wei Chow}{zju}
\icmlauthor{Shuicheng Yan}{sky}
\icmlauthor{Tat-Seng Chua}{nus}
\icmlauthor{Yueting Zhuang}{zju}
\icmlauthor{Hanwang Zhang}{sky,ntu}
\end{icmlauthorlist}

\icmlaffiliation{zju}{Zhejiang University}
\icmlaffiliation{nus}{National University of Singapore}
\icmlaffiliation{ntu}{Nanyang Technological University}
\icmlaffiliation{sky}{Skywork AI}

\icmlcorrespondingauthor{Juncheng Li}{junchengli@zju.edu.cn}

\icmlkeywords{Machine Learning, ICML}

\vskip 0.3in
]



\printAffiliationsAndNotice{}  

\begin{abstract}
For multimodal LLMs, the synergy of visual comprehension (textual output) and generation (visual output) presents an ongoing challenge. This is due to a conflicting objective: for comprehension, an MLLM needs to abstract the visuals; for generation, it needs to preserve the visuals as much as possible. Thus, the objective is a dilemma for visual-tokens. To resolve the conflict, we propose encoding images into \emph{morph-tokens} to serve a dual purpose: for comprehension, they act as visual prompts instructing MLLM to generate texts; for generation, they take on a different, non-conflicting role as complete visual-tokens for image reconstruction, where the missing visual cues are recovered by the MLLM. Extensive experiments show that morph-tokens can achieve a new SOTA for multimodal comprehension and generation simultaneously. Our project is available at \url{https://github.com/DCDmllm/MorphTokens}.
\end{abstract}

\section{Introduction}
\label{sec:intro}
State-of-the-art Multimodal Large Language Models (MLLMs) are still facing a great divide between visual comprehension (textual output) and generation (visual output). For comprehension tasks---``Tell me why the image [IMG] is funny''---we use GPT-4V~\cite{achiam2023gpt}; yet for generation---``Turn the image [IMG] into the style of Ghibli''---instead, we need DALL·E~\cite{ramesh2021zero}. Therefore, the community is interested in a \emph{unified, token-based, auto-regressive} MLLM framework~\cite{yu2023scaling}. Unlike traditional multimodal large models~\cite{wang2022image, rahman2020integrating}, this framework employs LLM as the core reasoning engine that drives both multimodal comprehension and generation.

\begin{figure}[t]
    \centering
    \includegraphics[width=1.0\linewidth]{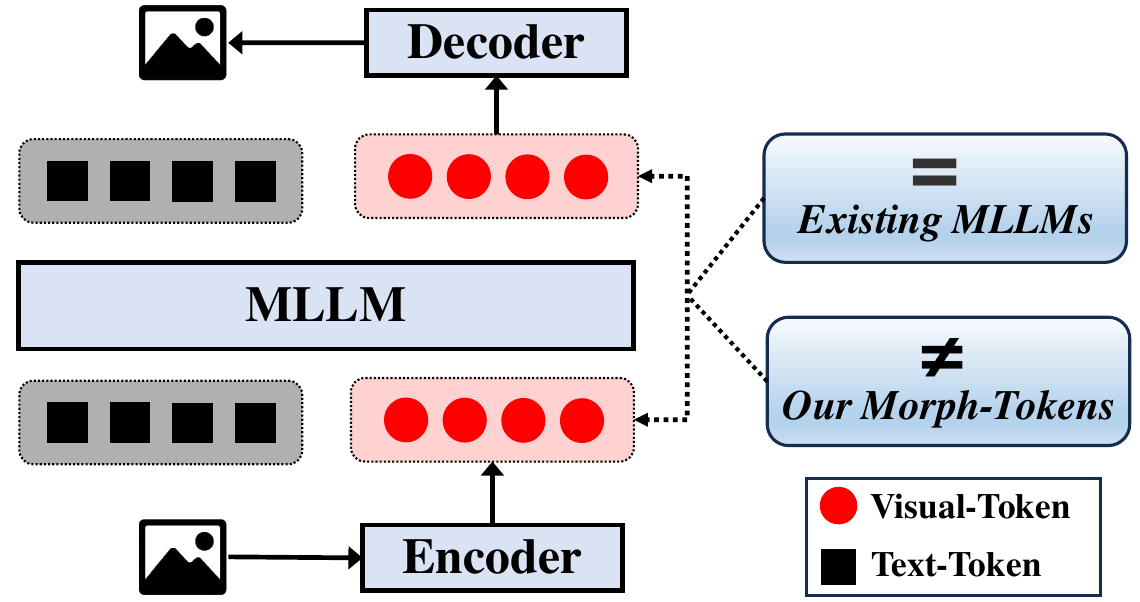}
    \vspace{-2em}
    \caption{Comparison between existing MLLMs and ours. The key difference is the equality between pre- and post-MLLM visual-tokens in training time.}
    \label{fig:training}
    \vspace{-1em}
\end{figure}

Existing solutions are straightforward. As shown in Figure~\ref{fig:training}, they have three steps: 1) images are encoded into visual-tokens by a tokenizer~\cite{fang2023eva, esser2021taming}; 2) these pre-MLLM visual-tokens are fed into an MLLM to complete vision-language comprehension tasks, where the MLLM is usually initialized from a pre-trained LLM, and 3) the post-MLLM visual-tokens are used to reconstruct input image in training or generate new images such as image editing in testing~\cite{koh2023generating, yu2023scaling}. More details of such MLLMs are reviewed in Section~\ref{sec:related}.

Yet, the synergy of comprehension and generation is not achieved. The primary challenge lies in the conflicting MLLM's training objectives for comprehension and generation tasks. Comprehension may discard visual features due to the need for visual abstraction, \textit{i.e.}, the MLLM training encourages the image tokenizer to output pre-MLLM visual-tokens invariant to task-irrelevant visual changes~\cite{instructblip, li2023finetuning}---a many-to-one map from images to tokens; conversely, generation requires preserving the visual details as much as possible, \textit{i.e.}, the post-MLLM visual-tokens should be equivariant to all the visual changes~\cite{wang2023equivariant}---a one-to-one map from tokens to images. The requirement for equality between pre- and post-MLLM visual-tokens poses a dilemma in the auto-regressive training of multimodal token sequences.

\begin{figure*}[t]
    \centering
    \includegraphics[width=1.0\linewidth]{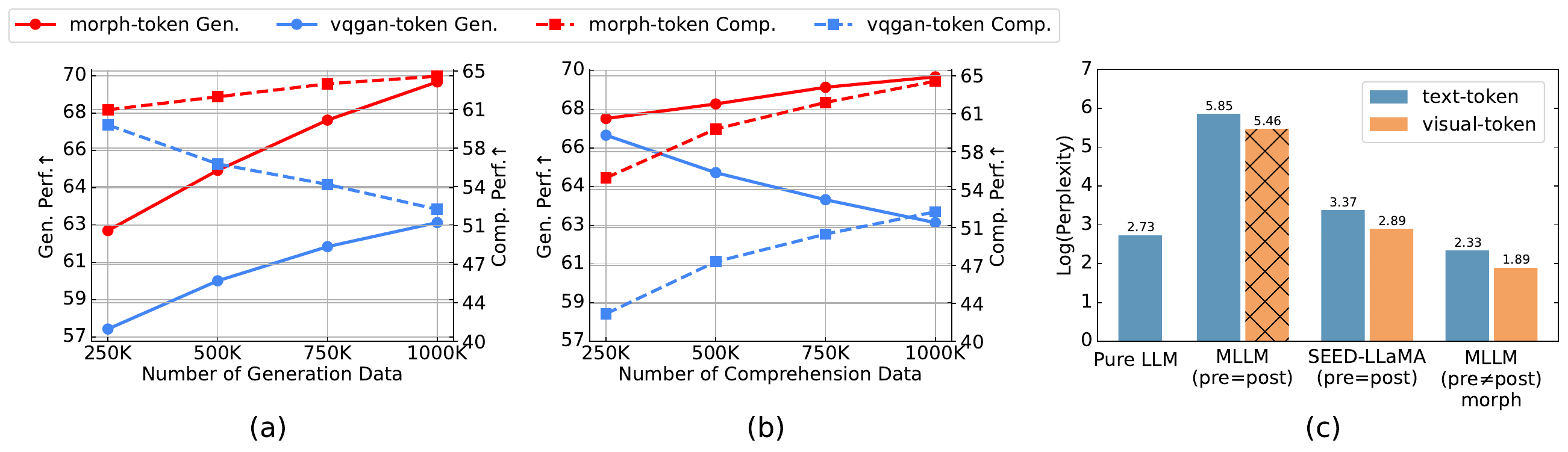}
    \vspace{-2em}
    \caption{(a, b) When the amount of visual comprehension (or generation) training data is fixed, the trend of comprehension \& generation performance with the number of visual generation (or comprehension) training data. (c) The average perplexity of the post-MLLM text-/visual-tokens with 1k text-image pair inputs. ``pre/post'' denotes pre-/post-MLLM.}
    \label{fig:trend_perplexity}
    \vspace{-1em}
\end{figure*}

As shown in Figure~\ref{fig:trend_perplexity}(a\&b), the comprehension (or generation) performance consistently decreases as the number of generation (or comprehension) training data increases, and vice versa. Another piece of evidence is shown in Figure~\ref{fig:trend_perplexity}(c), if we freeze an image tokenizer that encodes visually complete tokens capable of perfect image reconstruction (\textit{e.g.}, VQ-GAN~\cite{esser2021taming}), the multimodal auto-regressive training may even harm the original LLM's performance~\cite{kondratyuk2023videopoet, yu2022scaling}, that is, the perplexity of both the textual and visual-tokens is significantly increased.

We propose \textbf{Morph-Tokens} to resolve the conflict. As illustrated in Figure~\ref{fig:training}, the term ``morph'' implies a transformation where the pre-MLLM visual-tokens are \emph{not necessarily equal} to the post-MLLM ones. Specifically, the pre-MLLM tokens are abstract semantics, serving as visual prompts for comprehension tasks. In contrast, the post-MLLM tokens are visually complete tokens for image generation, thanks to the powerful comprehension ability of MLLM that recovers the lost visual features due to abstraction. We will detail the implementation of morph-token-based MLLM in Section~\ref{sec:mllm}. 

How do we use vision-language data to train morph-tokens to achieve the dual purpose without conflict? The key is to detach the textual and image reconstruction losses by using morph-tokens. This approach trains the MLLM to recognize the abstract pre-MLLM visual-tokens for comprehension. Simultaneously, it ensures their recovery back to visually complete tokens for image generation.  To this end, we propose a 3-stage training strategy (Section~\ref{sec:training}):

\begin{figure}[t]
    \centering
    \includegraphics[width=1.0\linewidth]{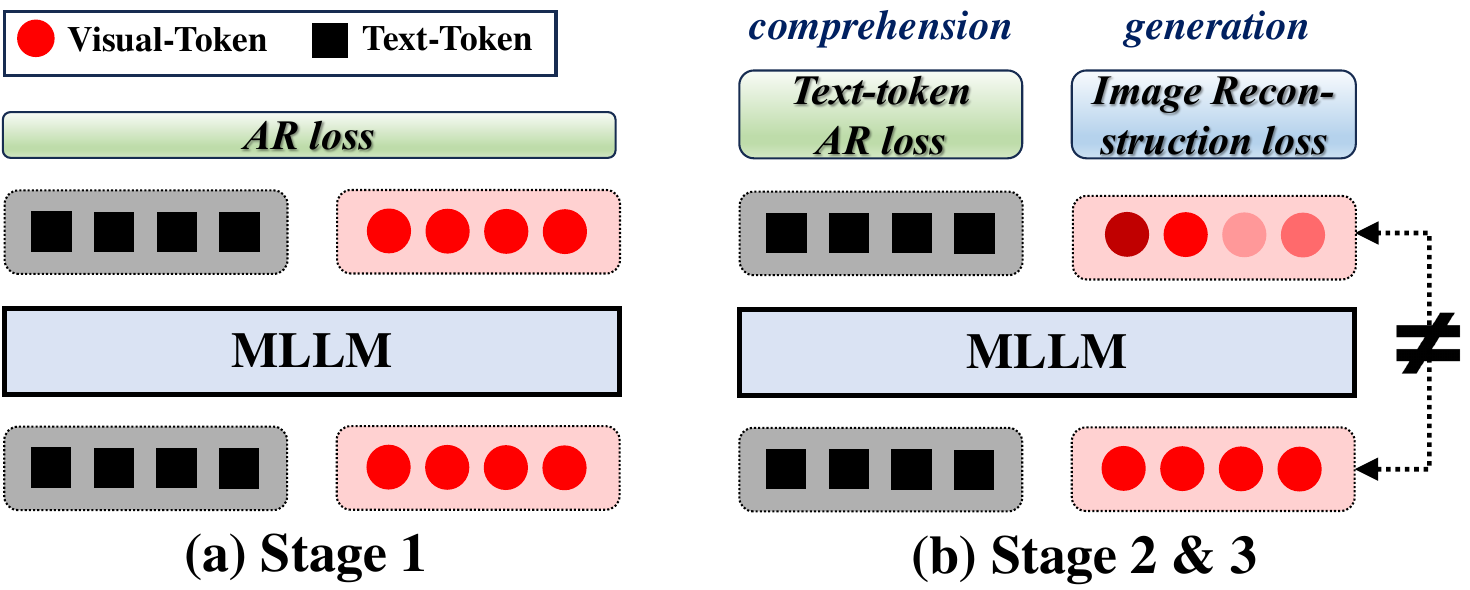}
    \vspace{-2em}
    \caption{Our 3-stage training strategy (Section~\ref{sec:training}). AR: auto-regressive.}
    \label{fig:stage}
    \vspace{-1em}
\end{figure}

\noindent\textbf{Stage 1}: As shown in Figure~\ref{fig:stage}(a), we use image-text pairs to train the morph-token encoder and the MLLM (initialized by an LLM) to auto-regress the concatenated morph-token and text-token sequence. This stage expands the token vocabulary, transitioning from LLM to MLLM. Note that although this stage requires equality between pre- and post-MLLM morph-tokens, there is no conflict due to the absence of a visual generation objective.

\noindent\textbf{Stage 2}: As shown in Figure~\ref{fig:stage}(b), we use the same image-text pairs to train the morph-token encoder, MLLM, and the decoder by both comprehension and generation tasks. For comprehension, \emph{i.e.}, image captioning, the pre-MLLM morph-tokens act as visual prompts instructing the MLLM to generate textual captions of the image; for generation, \textit{i.e.}, text-to-image generation, the post-MLLM morph-tokens play a different, non-conflicting role as visually complete tokens to reconstruct the input image. This stage can be viewed as an \emph{auto-encoding} process, unique in that it does not have a fixed morph-token bottleneck.

\noindent\textbf{Stage 3}: Similar to Stage 2, we use various vision-language tasks including both comprehension (e.g., VQA) and generation (e.g., image editing) to instruction-tune everything. 

Thanks to morph-tokens and the above training strategy, we observe a preliminary synergy shown in Figure~\ref{fig:trend_perplexity}. Through extensive experiments (Section~\ref{sec:exp}), besides a new SOTA on challenging vision-language benchmarks (\textit{e.g.}, DEMON~\cite{li2023finetuning}), we further find that our morph-token-based MLLM significantly outperforms others in multi-turn image editing (Figure~\ref{fig:icl_comp}) and in-context learning (Figure~\ref{fig:icl_gen}). Notably, these remarkable abilities to preserve image fidelity while understanding language instructions are rarely observed in prior works.

\section{Related Work}
\label{sec:related}

Thanks to Figure~\ref{fig:training}, we summarize relevant MLLMs into Table~\ref{tab:mllm}. We first divide them into two groups based on whether they use VQ-GAN~\cite{esser2021taming} or Stable Diffusion~\cite{rombach2022high} as the Decoder. Columns~2\&3 denote the nature of pre- and post-MLLM visual tokens, indicating whether they contain abstracted semantics or complete visuals. Columns~4\&5 assess the capability of these methods on multimodal in-context comprehension tasks and advanced image-editing tasks (\textsl{e.g.,} multi-turn editing with consistent image fidelity). The final column clarifies whether the methods detach the comprehension and generation losses. We can see that all the existing MLLMs require the equivalence between pre- and post-MLLM visual-tokens, which causes the conflict between comprehension and generation training objectives. Thus, none of them can achieve synergy on complicated comprehension and generation tasks. In contrast, we propose morph-tokens to detach the textual and image reconstruction losses, where the pre-MLLM visual-tokens are not necessarily equal to the post-MLLM ones (Column~2\&3), thus effectively resolving the conflict and achieving the synergy (Column~4\&5). More detailed comparisons are provided in Section~\ref{sec:exp}.

\begin{table}[t]
    \centering
    \renewcommand\arraystretch{0.3}
    \begin{adjustbox}{width=0.48\textwidth}
    \begin{tabular}{c|cc|cc|c}
    \toprule
     \multirow{2}{*}{Models}   & pre-MLLM  & post-MLLM & in-context & advanced  & detached\\
      & visual-token & visual-token & comp. & image-edit &loss\\
     \hline
    Cogview 
    & \multirow{2}{*}{complete} & \multirow{2}{*}{complete} 
    & \multirow{2}{*}{\ding{55}} & \multirow{2}{*}{\ding{55}} & \multirow{2}{*}{\ding{55}} \\
    \cite{ding2021cogview} & &  &  &&    \\   

    TEAL  & \multirow{2}{*}{complete} & \multirow{2}{*}{complete} & \multirow{2}{*}{\ding{55}} & \multirow{2}{*}{\ding{55}} & \multirow{2}{*}{\ding{55}}\\
    \cite{yang2023teal} & &  &  &&   \\   

    CM3Leon  & \multirow{2}{*}{complete} & \multirow{2}{*}{complete} & \multirow{2}{*}{\ding{55}} & \multirow{2}{*}{-} & \multirow{2}{*}{\ding{55}}\\
    \cite{yu2023scaling} & &  &  &&   \\   

     VideoPoet  & \multirow{2}{*}{complete} & \multirow{2}{*}{complete} & \multirow{2}{*}{\ding{55}} & \multirow{2}{*}{$\checkmark$} & \multirow{2}{*}{\ding{55}}\\
    \cite{kondratyuk2023videopoet} & &  &  &&   \\   
      \hline
    Gill  & \multirow{2}{*}{abstract} & \multirow{2}{*}{abstract} & \multirow{2}{*}{\ding{55}} & \multirow{2}{*}{\ding{55}} & \multirow{2}{*}{\ding{55}}\\
    \cite{koh2023generating} & &  &  &&  \\  

    Emu-I & \multirow{2}{*}{abstract} & \multirow{2}{*}{abstract} & \multirow{2}{*}{$\checkmark$} & \multirow{2}{*}{\ding{55}} & \multirow{2}{*}{\ding{55}}  \\
    \cite{sun2023generative} & &  &  &&   \\   

    Emu2-Chat  & \multirow{2}{*}{abstract} & \multirow{2}{*}{abstract} & \multirow{2}{*}{$\checkmark$} & \multirow{2}{*}{\ding{55}} & \multirow{2}{*}{\ding{55}} \\
    \cite{Emu2} & &  &  &&   \\ 
    
    DreamLLM  & \multirow{2}{*}{abstract} & \multirow{2}{*}{abstract} & \multirow{2}{*}{$\checkmark$} & \multirow{2}{*}{\ding{55}} & \multirow{2}{*}{\ding{55}}\\
    \cite{dong2023dreamllm} & &  &  &&  \\ 

    LaVIT  & \multirow{2}{*}{abstract} & \multirow{2}{*}{abstract} & \multirow{2}{*}{$\checkmark$} & \multirow{2}{*}{\ding{55}} & \multirow{2}{*}{\ding{55}}\\
    \cite{jin2023unified} & &  &  &&  \\ 
    
    Seed-LLaMA  & \multirow{2}{*}{abstract} & \multirow{2}{*}{abstract} & \multirow{2}{*}{$\checkmark$} & \multirow{2}{*}{\ding{55}} & \multirow{2}{*}{\ding{55}}\\
    \cite{ge2023planting} & &  &  &&    \\

    AnyGPT  & \multirow{2}{*}{abstract} & \multirow{2}{*}{abstract} & \multirow{2}{*}{$\checkmark$} & \multirow{2}{*}{\ding{55}} & \multirow{2}{*}{\ding{55}}\\
    \cite{zhan2024anygpt} & &  &  &&    \\

    Mini-Gemini  & \multirow{2}{*}{abstract} & \multirow{2}{*}{abstract} & \multirow{2}{*}{$\checkmark$} & \multirow{2}{*}{\ding{55}} & \multirow{2}{*}{\ding{55}}\\
    \cite{li2024mini} & &  &  &&    \\
      \hline
    \textbf{Morph-Token} (ours)  & abstract & complete & $\checkmark$ & $\checkmark$ & $\checkmark$ \\
    
    \bottomrule
    \end{tabular}
    \end{adjustbox}
    \vspace{-1em}
    \caption{\label{tab:mllm} Positioning of existing MLLMs and ours in terms pre-/post-MLLM visual-tokens, complex comprehension/generation capabilities, and if the conflicting training losses is detached.}
    \vspace{-0.5em}
\end{table}

\begin{figure*}[t]
    \centering
    \includegraphics[width=1.0\linewidth]{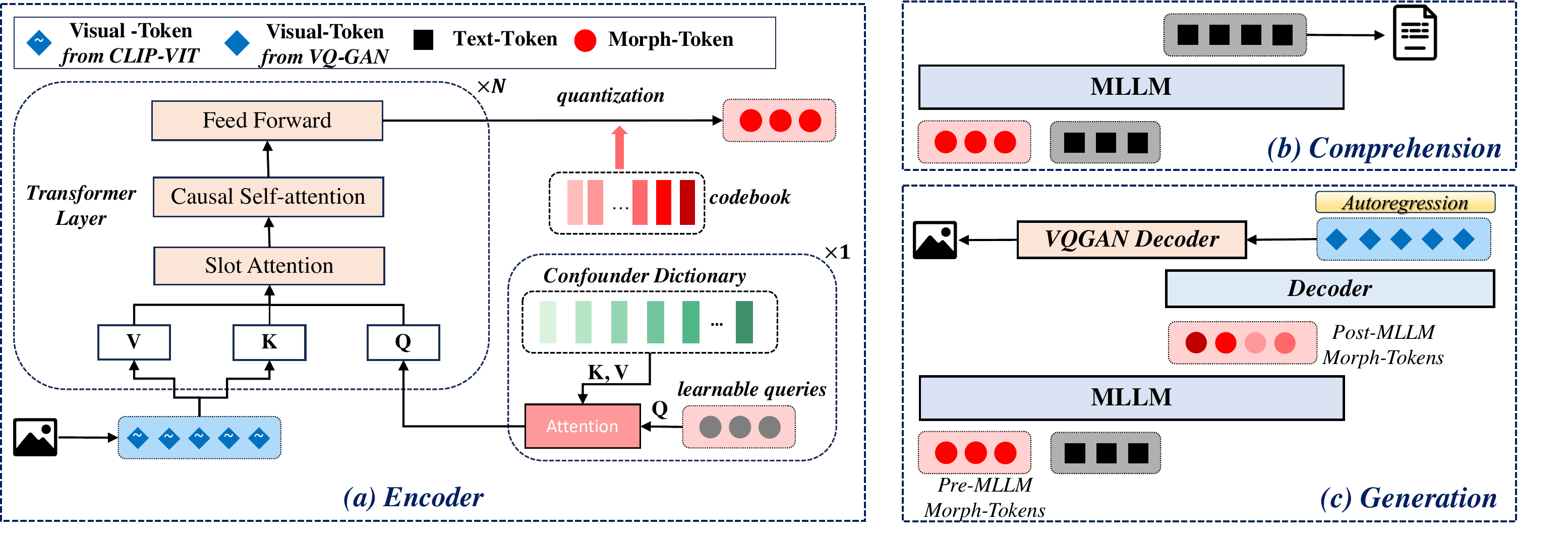}
    \vspace{-2em}
    \caption{(a) The encoder introduced in Section~\ref{sec:encoder}. (b) For comprehension tasks, pre-MLLM morph-tokens instruct MLLM to generate texts. (c) For generation tasks, post-MLLM morph-tokens are first decoded into a lower-level visual-tokens (blue), introduced in Section~\ref{sec:mllm}. Then, they are fed into the VQ-GAN decoder to generate high-fidelity images.}
    \label{fig:main}
    \vspace{-1em}
\end{figure*}

\section{Method}
We introduce the proposed morph-token-based MLLM in Figure~\ref{fig:main}. The detailed implementations are in Appendix~\ref{sec:B}.
\subsection{Encoder}
\label{sec:encoder}
As illustrated in Figure~\ref{fig:main}(a), given visual-tokens $\mathcal{V}$ extracted from an image, \textit{i.e.}, by CLIP-ViT~\cite{fang2023eva}, our encoder is proposed to abstract these visuals by transforming them into morph-tokens $\mathcal{M}$, which serve as visual prompts for comprehension tasks. As shown in Figure~\ref{fig:main}(a), we use Q-former~\cite{li2023blip} to abstract $\mathcal{V}$ into token embeddings, which are quantized into discrete morph-tokens $\mathcal{M}$:
\begin{equation}
\label{eq:aggregation}
\small
\begin{aligned}
    \mathcal{M} &= \texttt{Quantizer}(\ \texttt{Qformer}(Q = \mathcal{Q}, K = \mathcal{V}, V = \mathcal{V})\ )
\end{aligned}
\end{equation}
where the arguments ($Q$, $K$, $V$) denote the query, key, and value of Q-former, $\mathcal{Q}$ is a set of query embeddings obtained from below.

As the CLIP-ViT visual tokens are merely ﬂattened 2D patch features, the attention across such spatial visual-tokens is usually spuriously correlated~\cite{wang2020visual}. To remove such spatial confounding effect, inspired by~\cite{wang2020visual}, we integrate a deconfounder dictionary $\mathcal{D}$ to initialize the above query embeddings $\mathcal{Q}$, making the resultant morph-tokens behave more like natural language that is causal~\cite{pearl2018book}. Specifically, given a set of learnable query vectors $\mathcal{G}$, we initialize $\mathcal{D}$ as a learned dictionary of pre-trained ViT-VQGAN and adopt a single-layer Q-former to obtain $\mathcal{Q}$:
\begin{equation}
\label{eq:aggregation2}
\small
\begin{aligned}
    \mathcal{Q} = \texttt{Single-Qformer}(Q = \mathcal{G}, K = \mathcal{D}, V = \mathcal{D}).
\end{aligned}
\end{equation}

\subsection{Morph-token-based MLLM}
\label{sec:mllm}
After transforming $\mathcal{V}$ into $\mathcal{M}$, the morph-token-based MLLM, where the pre-MLLM morph-tokens $\mathcal{M}$  are not necessarily equal to the post-MLLM ones $\hat{\mathcal{M}}$.
Specifically, as shown in Figure~\ref{fig:main} (b\&c), for comprehension tasks, $\mathcal{M}$ serves as visual prompts to instruct the MLLM to generate text-tokens $\mathcal{Y}$; for generation tasks, MLLM produces another set of post-MLLM morph-tokens $\hat{\mathcal{M}}$ which recover the visual features lost by $\mathcal{M}$, thus $\hat{\mathcal{M}}$ can generate images.
In this way, $\mathcal{M}$ and $\hat{\mathcal{M}}$ effectively resolve the conflicting objectives of visual comprehension and generation. However, $\hat{\mathcal{M}}$ \emph{per se} cannot yet generate images of high fidelity because it is not reasonable to force the MLLM to recover all the high-frequency visual details. Therefore, we need to further decode $\hat{\mathcal{M}}$ into lower-level visual-tokens $\mathcal{X}$ that can be finally decoded back to pixels by VQGAN. This decoder is introduced below.

\subsection{Decoder}
\label{sec:decoder}
Our design philosophy is to allow the lower-level visual tokens to auto-regressively generate their own visual distributions. This design aims to disentangle the different distributions of natural language and visual tokens (Figure~\ref{fig:trend_perplexity}(c)). Such disentanglement further helps in resolving the conflict during MLLM training. As illustrated in Figure~\ref{fig:main}(c),
\begin{equation}
\label{eq:3}
\small
\begin{aligned}
\textrm{Image} = \texttt{VQGAN} (\mathcal{X}), ~~  \mathcal{X} = \texttt{Decoder}(\hat{\mathcal{M}}), 
\end{aligned}
\end{equation}
where $\hat{\mathcal{M}}$ serves as a higher-level visual prompt to instruct \texttt{Decoder}, a decoder-only Transformer that decodes $\mathcal{X}$ which can be fed into a pre-trained VQGAN decoder to generate an image.

\subsection{Training Strategy}
\label{sec:training}
We detail the 3-stage strategy, as illustrated in Figure~\ref{fig:stage}, for training the morph-token-based MLLM.

\noindent\textbf{Stage 1: Initialization.}
We aim to extend the token vocabulary of a pre-trained Vicuna~\cite{vicuna}, transitioning it from LLM to MLLM.
We use $\sim$30M image-text pairs, which are concatenated in two formats, \textit{i.e.}, $\langle \mathcal{M}, \mathcal{Y}\rangle$ and $\langle \mathcal{Y}, \mathcal{M}\rangle$.
As shown in Figure~\ref{fig:training}(a), we fine-tune the morph-token encoder and the LLM to maximize the auto-regressive likelihood of the two-format concatenated tokens:
\begin{equation}
\small
\begin{aligned}
    \underset{\theta_{Enc},\  \theta_{LLM}}{{\arg\max}} \, \text{log} P\left( t_i\in\{\mathcal{M}, \mathcal{Y}\}\  |\  t_{<i}\in\{\mathcal{M}, \mathcal{Y}\}\right)
\end{aligned}
\end{equation}
where $t_i$ denotes a morph-/text-token, $\theta_{Enc}$ is the encoder parameters, and $\theta_{LLM}$ denotes the LoRA~\cite{hu2021lora} parameters attached to Vicuna. In particular, we set the token length: $|\mathcal{M}| = 32$ and $|\mathcal{Y}| = 512$.
Recall that although this stage requires equality between pre- and post-MLLM morph-tokens, there is no conflict due to the absence of a visual generation objective. The resultant vocabulary size of our MLLM is 8,192 morph-tokens and 32,000 text-tokens.

\noindent\textbf{Stage 2: Auto-encoding Morph-Tokens.}
We use the same image-text pairs. As shown in Figure~\ref{fig:stage}(b), for image-captioning comprehension task, we use the token format $\langle \mathcal{M}, \mathcal{Y}\rangle$, where $\mathcal{M}$ serves as a visual prompt to instruct the MLLM to generate $\mathcal{Y}$ auto-regressively:
\begin{equation}
\label{eq:5}
\small
\begin{aligned}
    \underset{\theta_{Enc},\  \theta_{LLM}}{{\arg\max}} \, \text{log} P\left( t_i\in \mathcal{Y}\  |\  t_{<i}\in\{\mathcal{M}, \mathcal{Y}\}\right).
\end{aligned}
\end{equation} 
For text-to-image generation task, we use the token format $\langle \mathcal{Y}, \mathcal{M}\rangle$ to feed into MLLM that generate $\hat{\mathcal{M}}$ auto-regressively. Then, $\hat{\mathcal{M}}$ is decoded into $\mathcal{X}$ by Eq.~\eqref{eq:3} auto-regressively:
\begin{equation}
\label{eq:6}
\small
\begin{aligned}
    \underset{\theta_{Enc},\  \theta_{LLM},\  \theta_{Dec}}{{\arg\max}} \, \text{log} P\left( x_i\in \mathcal{X}\  |\  x_{<i}\in \mathcal{X}, \{\hat{m}\}_{j=1}^N \in \hat{\mathcal{M}}\right)
\end{aligned}
\end{equation}
where $\theta_{Dec}$ denotes the parameters of the decoder. Recall that the above reconstruction loss does not impose $\hat{\mathcal{M}} =  \mathcal{M}$. Thus, the training objectives of Eq.~\eqref{eq:5} and Eq.~\eqref{eq:6} do not conflict. During inference, if the generated $|\hat{\mathcal{M}}|<|\mathcal{M}|$, we complete it with special $\langle Emp\rangle$ tokens; if $|\hat{\mathcal{M}}|>|\mathcal{M}|$, we trim it to $|\mathcal{M}|$.
This stage can be viewed as an auto-encoding process: $\textrm{Image}\!\!\to\!\!\mathcal{V}\!\!\to\!\!\mathcal{M}\!\!\to\!\!\hat{\mathcal{M}}\!\!\to\!\!\mathcal{X}\!\!\to\!\!\textrm{Image}$, unique in that it does not have a fixed morph-token bottleneck.

\noindent\textbf{Stage 3: Instruction Tuning.}
Besides the image-text pairs used in the above two stages, we use extensive interleaved image-text data as ``\texttt{<Instructions>}'' to enhance the MLLM's comprehension and generation capabilities in complex scenarios, \textit{e.g.}, \texttt{<Instructions>}$=\langle\mathcal{M}, \mathcal{Y}, \mathcal{M}\rangle$. We use 
the instruction format ``\texttt{USER: <Instructions> ASSISTANT: <Answers>.}'', where ``\texttt{USER:}'' and  ``\texttt{ASSISTANT:}'' are prompt tokens, and the training loss is as the same as Stage 2. See Appendix~\ref{sec:C} for the instruction examples of diverse tasks.

\section{Experiments}
\label{sec:exp}
\begin{table*}[t]
    \centering
    \begin{adjustbox}{width=\textwidth}
    \begin{tabular}{l|cc|cc|cccc|cc|cc}
    \toprule
    \multirow{2}{*}{Models} & \multirow{2}{*}{Size} & Image & \multicolumn{2}{c|}{Image caption} &  \multicolumn{4}{c|}{Image QA} &  \multicolumn{2}{c|}{Video QA} &  \multicolumn{2}{c}{MME Bench} \\
     && Gen                                       &NoCaps     &Flickr30K	 & GQA & VSR & ICONQA &	HM & MSVDQA	& MSRVTTQA& Perception & Cognition  \\
    \hline
    Flamingo     & 9B    & \ding{55}    & -         & 61.5  & -     & -     & -     & 57.0    & 30.2  & 13.7 & -         & -   \\
    BLIP-2                                 & 11B    & \ding{55}    & 98.4     & 73.7  &44.6   &68.2   &45.4   &52.0   &34.4    &17.4 &1293.8   &290.0    \\
    InstructBlip                          & 11B    & \ding{55}    & 120.0     & 83.5  & 47.9  & 65.6  & \textbf{51.2}  & 54.1  & 44.3   &25.6 & 1212.8  & 291.8  \\  
    MiniGPT4                              & 13B    & \ding{55}    & -         & -     &30.8      &41.6      &37.6      &-      &-          &-      &581.7      &144.3      \\
    LLaVA                                   & 13B   & \ding{55}    & -         & -     &41.3   &51.2   &43.0   &-      & -    &- &\textbf{1531.3}      &295.4      \\
    mPlug-Owl                               & 13B    & \ding{55}  &-   & 85.1         & 56.1     &-      &-      &-      &42.4      &23.6          &1450.2      &313.2           \\ \hline
    Emu-I                                   & 13B   &-  & 106.8         & 80.9         & 46.0     & 53.9      &42.9      &56.2      &37.0      &21.2          &660.9      &257.9            \\
    Emu2-Chat                               & 37B    & -         & 119.8         & 86.0     &\textbf{65.1}      &50.4      &49.9      &57.2      &49.0          &31.4      &1315.7      &307.5      \\
    Seed-LLaMA                              & 8B    &$\checkmark$ & 90.4         & 66.7     &34.8      &45.2      &36.0      &50.4      &45.2          &35.3      &736.7      &235.5      \\
    LaVIT                                   & 7B    &$\checkmark$ & 114.2         & 83.0     & 46.8      &60.4      &36.8      &53.0      &-          &-      &997.9     &240.7      \\ 
    \hline
    \textbf{Ours}                           & 8B    &$\checkmark$ & \textbf{124.0}         & \textbf{87.5}     &56.8      &\textbf{69.8}      &47.6      &\textbf{62.0}      &\textbf{50.9}          &\textbf{37.2}      &\underline{1477.7}      &\textbf{389.3}      \\
    \hline
    \end{tabular}
    \end{adjustbox}
    \vspace{-1em}
    \caption{\label{tab:main_comp} Comparison for multimodal comprehension. ``Image Gen'' denotes whether the model can generate images besides texts (Emu-I and Emu2-Chat only possess the image generation capability in the versions prior to instruction-tuning).}
\end{table*}
\definecolor{COLOR_MEAN}{HTML}{f0f0f0}

\begin{table}[t]
    \centering
    \begin{adjustbox}{width=0.5\textwidth}
    \begin{tabular}{l|ccccccc}
    \toprule

    Models &   MMD & VST & VRI & MMC & KGQA & TRQA & MMR  \\
    \hline
    Flagmingo    &   16.9   &   24.2   &   13.9   &   21.7  &  32.0  &   30.6  &  41.6 \\ 
    Blip-2   &   26.1  &   21.3   &   10.7   &   17.9  &  39.2  &   33.5  &  39.7 \\ 
    InstructBlip    &   33.6   &   24.4   &   11.5   &   21.2  &  47.4  &   44.4  &  48.6 \\ 
    MiniGPT-4    &   13.7   &   17.1   &   8.0   &   16.6  &  30.3  &   26.4  &  43.5 \\ 
    LLaVA   &   7.8   &   10.7   &   8.3   &  15.9  &  36.2  &   28.3  &  41.5 \\ 
    mPlug-Owl  &   12.7   &   19.3   &   5.4   &   16.3  &  33.3  &   32.5  &  42.5 \\ 
    VPG-C  &   \textbf{37.5}  &   25.2   &  25.9   &  22.2  &  48.6  &  44.9  &  50.3 \\ 
    \hline
    Emu-I  &   25.6   &   16.1   &   13.4   &   23.1  &  46.4  &   32.2  &  42.6 \\ 
    Emu2-Chat  &   26.8   &   19.8   &   13.6   &   19.3  &  54.6  &   44.2  &  46.7 \\ 
    Seed-LLaMA  &   10.4   &   15.7   &   11.5   &   18.5  &  30.9  &   33.3  &  44.6 \\ 
    LaVIT  &   36.5   &   25.5   &   10.8   &   26.7  &  38.0  &   38.2  &  45.8 \\  
    \hline
    Ours   &   32.2   &   \textbf{27.4}   &   \textbf{27.4}   &   \textbf{28.0}  &  \textbf{56.4}  &   \textbf{47.7}  &  \textbf{54.9} \\ 
    \bottomrule
    \end{tabular}
    \end{adjustbox}
    \vspace{-1em}
    \caption{\label{tab:demon}Average results of zero-shot evaluation on each task category of DEMON Benchmark.}
    \vspace{-1.2em}
\end{table}
Through instruction-tuning on extensive supervised image-text data from diverse tasks (e.g.,  text-to-image generation, image editing, image QA, multi-image understanding), our model evolves into a versatile multimodal generalist, excelling in zero-shot vision-language comprehension and synthesis tasks. 
We conduct thorough experiments across a wide range of vision-language tasks, 
making comparisons primarily with MLLMs~\cite{koh2023generating, sun2023generative, ge2023planting, jin2023unified} designed for both visual comprehension and generation.
And we also compare with some widely-used MLLMs~\cite{alayrac2022flamingo, li2023blip, instructblip, zhu2023minigpt, liu2023visual, ye2023mplug} that specialize solely in comprehension tasks. 
For detailed experimental setups and implementation specifics, please refer to Appendix~\ref{sec:C}.

\subsection{Zero-shot Multimodal Comprehension}
\paragraph{Image Caption and VQA.} 
We first evaluate our model on a wide range of academic benchmarks including image captioning and image/video question answering datasets. 
As shown in Table~\ref{tab:main_comp}, our model achieves competitive performance in both image and video understanding tasks.
Specifically, 
(1) Compared to specialized visual comprehension MLLMs like  InstructBlip~\cite{instructblip},
as well as models like Emu2-Chat that  primarily concentrate on  comprehension tasks with significantly larger parameter scales, 
our method simultaneously maintains the image generation capabilities and achieves enhanced multimodal comprehension capabilities.
For instance, our model consistently attains higher Cider scores in image captioning tasks and performance in Video QA tasks, which requires the understanding of multiple images. 
(2) Furthermore, Compared to previous SOTA models (Seed-LLaMA and LaVIT) which are capable of both visual generation and understanding, our model consistently exhibits stronger performance in all these image captioning and Visual Question Answering benchmarks, which underscores the robust multimodal comprehension capacity of our model.

\paragraph{MLLM-oriented Comprehension Benchmarks.}
Our zero-shot evaluation also encompasses recent MLLM-oriented comprehension benchmarks, including MME~\cite{fu2023mme} and DEMON benchmark~\cite{li2023finetuning}. We have the following observations:
(1) The results on MME in Table~\ref{tab:main_comp} underscore the strong generalizability of our model to follow a diverse range of single-image instructions. Especially when compared to similar MLLMs with certain image generation capabilities, our model demonstrates a significant advantage in both perception and cognition abilities.
(2) Table~\ref{tab:demon} showcases the superior performance of our model on the DEMON benchmark, which is specifically designed to evaluate a model's capability of in-context learning on following demonstrative instructions. 
And our model outperforms the previous SOTA model in the DEMON benchmark, i.e., VPG-C~\cite{li2023finetuning}, across the majority of task categories. For instance, we achieve performance improvements of 5.6\% in multi-modal cloze (MMC) tasks and 7.8\% in knowledge-grounded image QA (KGQA) tasks compared to VPG-C, which underscores our advanced ability to associate interleaved text-image inputs for stronger in-context understanding.

\subsection{Zero-shot Image Synthesis}
\paragraph{Text-to-Image Generation.}
To evaluate our model's capabilities in zero-shot image synthesis, we first evaluate the text-to-image generation on MS-COCO\cite{lin2014microsoft} (30K randomly sampled data from validation set, and 5K data in karpathy test set) and Flickr30K~\cite{young-etal-2014-image} (1K data in test set), and compute pair-wise CLIP similarity score as the evaluation metric following previous works~\cite{koh2023generating, ge2023planting}. 
As shown in Table~\ref{tab:t2i}, compared to all other existing MLLMs, the images generated from textual descriptions by our method consistently exhibit higher similarity with the ground-truth images. 
This highlights that our model facilitates better vision-text alignment, thereby effectively transforming text prompts into more relevant images.

\begin{table}[t]
    \centering
    \begin{adjustbox}{width=0.38\textwidth}
    \begin{tabular}{lccc}
        \toprule
        \multirow{2}{*}{Models} & COCO & COCO & Flickr \\
        & \textit{Karpathy test} & \textit{val30k} & \textit{test} \\ \midrule
        GILL & 68.4 & 67.5 & 65.2 \\
        Emu & 65.6 & 66.5 & 64.8 \\
        Emu2-Gen & 68.6 & 67.6 & 64.9\\
        SEED-LLaMA & 68.2 & 70.7 & 65.6\\
        LaVIT & - & 68.4 & 63.5 \\ \hline
        Ours & \textbf{70.6} & \textbf{72.2} & \textbf{68.8} \\
        \bottomrule
    \end{tabular}
    \end{adjustbox}
    \vspace{-0.6em}
    \caption{\label{tab:t2i} Zero-shot Evaluation of text-to-image generation.}
    \vspace{-1em}
\end{table}

\begin{table}[t]
    \centering
    \begin{adjustbox}{width=0.48\textwidth}
    \begin{tabular}{lcccccccc}
    \toprule
    ~  & \multicolumn{2}{c}{\textbf{EVR}} & ~ & \multicolumn{2}{c}{\textbf{MA5K}} & ~ & \multicolumn{2}{c}{\textbf{MagicBrush}} \\
    \cmidrule{2-3} \cmidrule{5-6} \cmidrule{8-9} Method  & L1$\downarrow$ & CVS$\uparrow$  & ~ & L1$\downarrow$ & LPIPS$\downarrow$  & ~ & L1$\downarrow$ & CVS$\uparrow$  \\ \hline
    InsPix2Pix      & 18.9     &  81.4    & ~     &  17.6   &35.9  & ~     & 10.1    &85.2 \\
    LGIE            & 15.9     &82.0     & ~     & 14.4    & 32.7  & ~     & 8.4     & 88.9 \\
    MGIE     &    16.3 &  81.7    & ~     & \textbf{13.3}     &  29.8 & ~     & 8.2     &  \textbf{91.1}\\ \hline
    Gill & 31.8 & 65.0 & ~ & 27.4 & 44.3  & ~& 28.3 & 75.2 \\
    Emu            & 30.7    & 69.2     & ~     & 27.2     &43.2  & ~     & 27.9     & 78.5 \\
    Emu2-gen     & 22.8     & 80.3     & ~     & 20.5     &28.4  & ~     & 19.9     & 85.7 \\
    SEED-LLaMA        & 28.4     & 72.3     & ~     & 24.6     &39.0  & ~     & 24.5     & 80.9 \\
    LaVIT   & 26.8     & 73.8     & ~     & 25.1     & 36.9  & ~     & 25.3     & 81.1 \\
    Ours        & \underline{\textbf{15.3}}     & \underline{\textbf{82.6}}     & ~     & \underline{14.6}    & \underline{\textbf{27.9}}  & ~     & \underline{\textbf{7.6}}     & \underline{87.9} \\
   
    \bottomrule
    \end{tabular}
    \end{adjustbox}
    \vspace{-1em}
    \caption{\label{tab:edit} Zero-shot image editing results. The first three rows consist of models specialized in image editing. The best results among  MLLMs capable of both multimodal comprehension and generation are \underline{underline}.}
    \vspace{-1em}
\end{table}

\begin{figure}[t]
    \centering
    \includegraphics[width=\linewidth]{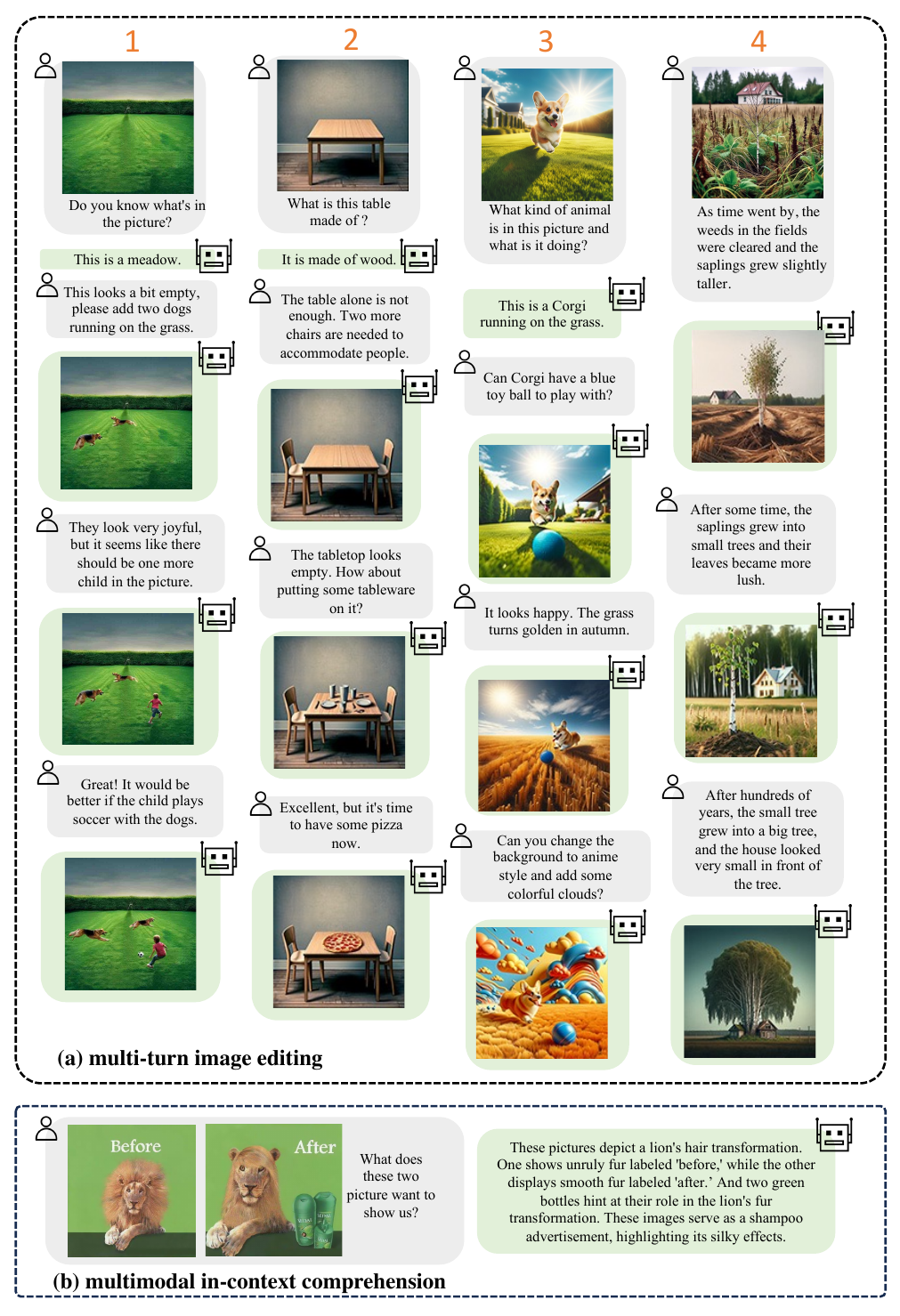}
    \vspace{-1.5em}
    \caption{ Qualitative results on multi-turn image editing and multimodal in-context comprehension.}
    \vspace{-1em}
    \label{fig:icl_comp}
    \vspace{-1em}
\end{figure}

\begin{figure*}[t]
    \centering
    \includegraphics[width=1.0\linewidth]{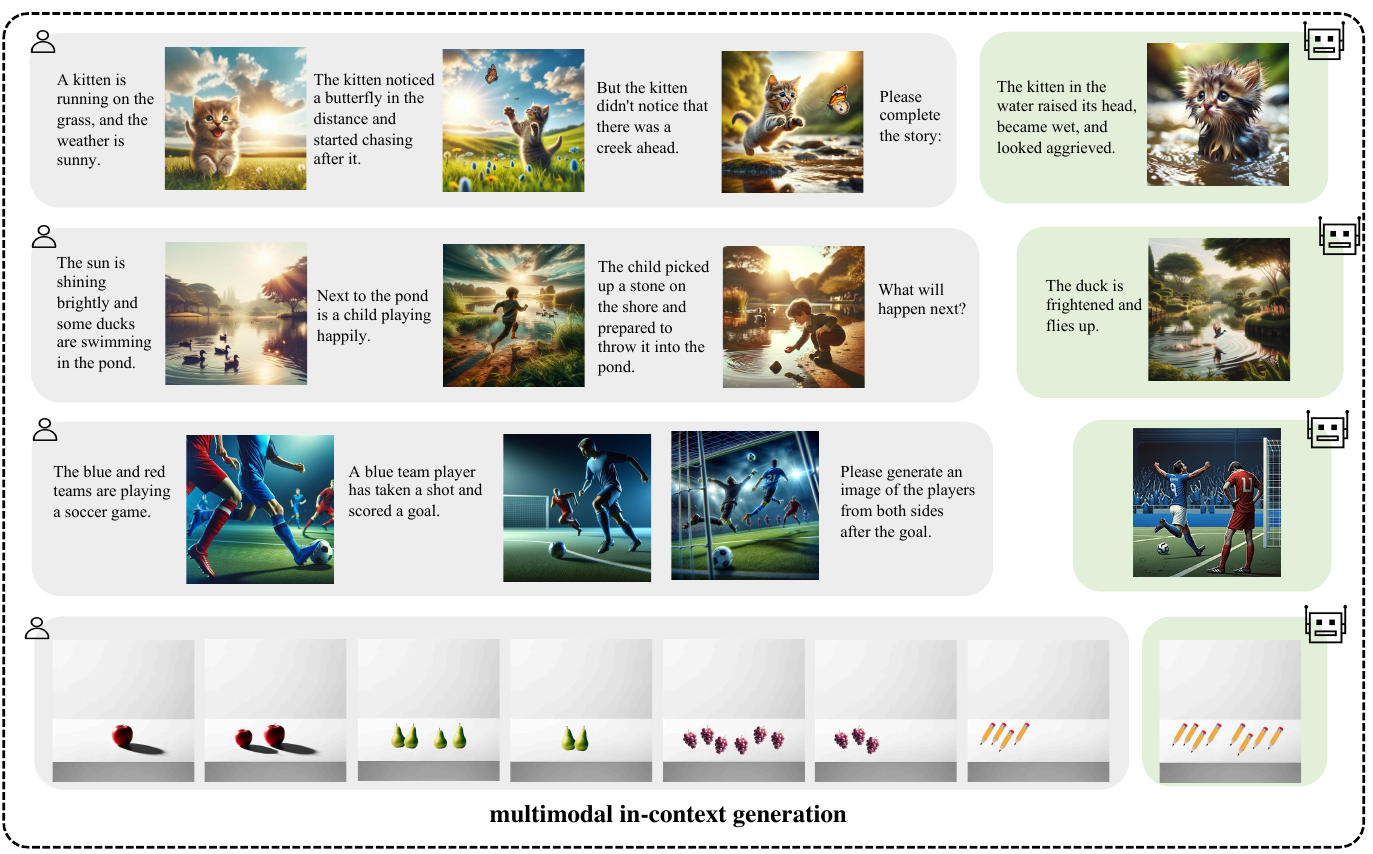}
    \vspace{-2em}
    \caption{ Qualitative results on multimodal in-context generation.}
    \vspace{-0.5em}
    \label{fig:icl_gen}
\end{figure*}

\paragraph{Image Editing.}
We further evaluate our model in zero-shot instruction-based image editing across three datasets: EVR~\cite{tan2019expressing}, MA5k~\cite{shi2021learning}, and MagicBrush~\cite{zhang2023magicbrush}. 
Following \citet{fu2023guiding}, for EVR and MagicBrush, we treat the standard pixel difference (L1) and visual feature similarity from the CLIP visual encoder (CVS) between generated images and ground-truth goals as the evaluation metrics. For MA5K, we utilize L1 and Learned Perceptual Image Patch Similarity (LPIPS~\citealp{zhang2018unreasonable}) as the evaluation metrics.
The experimental results are shown in Table~\ref{tab:edit}, leading to the following observations:

\textbf{First}, Across all datasets and metrics in image editing, our model significantly outperforms existing Multimodal MLLMs that possess unified multimodal comprehension and generation capabilities.  
Moreover, when compared to image-editing specialists (e.g., InsPix2Pix~\citealp{brooks2023instructpix2pix}, MGIE~\citealp{fu2023guiding}), our model still achieves stronger performance. 
It consistently surpasses InsPix2Pix and shows greater efficacy than previous SOTA specialists, i.e.,  MGIE across most datasets. 
For instance, on the EVR dataset, our approach achieves lower L1 scores and higher CLIP similarity compared to MGIE.
\textbf{Second}, compared with other MLLMs such as Emu2-Gen, our model demonstrates a more pronounced superiority in terms of the L1 score than it does on the CVS metric.
We find that this disparity is attributed to the inability of existing MLLMs to maintain image fidelity throughout the editing process.
Therefore, while models like Emu2 still achieve decent CVS scores, the performance in terms of L1 scores is not ideal. 
In contrast, our model not only effectively comprehends instructions to execute image editing, but also well preserves image fidelity, which leads to a significant advantage in L1 scores over other MLLMs.

\subsection{Emergent Abilities}
\paragraph{Multi-turn Image Editing.} As shown in Figure~\ref{fig:icl_comp}.a, in multi-turn image editing scenarios, our model is capable of effectively interpreting diverse user instructions to edit the image, while ensuring the preservation of image fidelity. Visual objects not intended for alteration retain a high level of consistency before and after editing.

\paragraph{Mulmodal In-context Learning.} 
Our model also exhibits an advanced capability for multimodal in-context learning. 
In comprehension tasks (Figure~\ref{fig:icl_comp}.b), it can keenly identify the connections between input images and provides insightful analyses. 
Moreover, in generation tasks (Figure~\ref{fig:icl_gen}), it can effectively understand the complete meaning of interleaved image-text inputs and engages in compositional image generation following instructions. 
Even in the absence of any natural language instructions, merely provided with several image pairs, our model is capable of inferring the patterns of change between images to understand the task requirements, thereby generating the desired image.

\subsection{In-Depth Analysis}

\paragraph{Impact of ``Morph'' .} 

To explore the impact of morph tokens, we eliminate the design of ``morph'', and train the following ablation models which require the equivalence between pre- and post-MLLM visual tokens:
\textbf{(1) detail-detail}: following TEAL~\cite{yang2023teal}, both pre- and post-MLLM visual tokens contain complete visuals. 
\textbf{(2 \& 3) abstr-abstr (SD \& VQ)}: both pre- and post-MLLM visual tokens contain abstracted semantics. And image generation is facilitated either through the use of SD models (following SEED-LLaMA) or via another decoder-only transformer to reconstruct the detailed visuals that can be converted into the image with a VQ-GAN decoder.
The implementation details are shown in Appendix~\ref{sec:C}.
We evaluate their performance on Demon benchmark and image-editing tasks (with L1 as the evaluation metric), as shown in Table~\ref{tab:abla-demon} and \ref{tab:abla-edit}.

\textbf{Firstly}, the results of Row1 show that when pre-MLLM tokens contain detailed visual semantics, MLLMs not only fail to facilitate effective visual comprehension (as evidenced by unsatisfactory performance on Demon), but also fall short in image editing, which necessitates a comprehensive understanding of the image being edited.
\textbf{Secondly}, The results of Row2 and Row3 show that when both pre- and post-MLLM visual tokens embody abstract semantics, MLLMs firstly struggle to effectively conduct image editing tasks without the support of visually-complete post-MLLM tokens.
Moreover, forcing them to possess visual generation capabilities also notably compromises their effectiveness in executing complex visual comprehension tasks like DEMON.

\begin{table}[t]
    \centering
    \begin{adjustbox}{width=0.44\textwidth}
    \begin{tabular}{l|ccccccc}
    \toprule
    \multirow{2}{*}{Model}
     & \multicolumn{3}{c}{Image $\rightarrow$ Text} & \multicolumn{3}{c}{Text $\rightarrow$ Image}  \\
    \cmidrule(l){2-4}\cmidrule(l){5-7} & R@1        & R@5      & R@10      & R@1       & R@5       & R@10   &R@m   \\ \hline
    BLIP-2     & 81.9& 98.4     & 99.7      & 82.4      & 96.5      & 98.4     &92.9     \\
    SEED &\textbf{91.0}	&99.5	&\textbf{100.0}	&79.3	&94.8	&97.1	&93.6\\
    LaVIT & 	83.0 &99.2	&99.7	&78.3	&96.2	&97.5	&92.3	\\ \hline
    Ours  &88.8	&\textbf{99.7}	&\textbf{100.0}	&\textbf{85.2}	&\textbf{97.1}	&\textbf{98.7}	&\textbf{95.0}	\\

    \bottomrule
    \end{tabular}
    \end{adjustbox}
    \vspace{-0.5em}
    \caption{\label{tab:retrieval}Evaluation of Image-Text Retrieval on Flickr30K.}
    \vspace{-1.5em}
\end{table}

\paragraph{Effectiveness of Individual Components.}
To further investigate the effectiveness of individual components, we train the following ablation models and also evaluate on Demon benchmark and image-editing tasks: \textbf{(1) w/o decoder}: we force the MLLM to recover all the high-frequency visual details, where post-MLLM visual tokens are directly fed into a VQ-GAN decoder for image generating. The results of Row4 in Table~\ref{tab:abla-demon} and \ref{tab:abla-edit} show that it is quite difficult for MLLM to directly autoregress such lower-level visual-tokens that can be finally decoded back to pixels by VQ-GAN. And an additional decoder is essential to alleviate its burden in recovering the lost visual features.
\textbf{(2) w/o deconfounded}: we remove the deconfounding design in the encoder. The results of Row 5 in Table~\ref{tab:abla-demon} and \ref{tab:abla-edit} demonstrate that deconfounding enables visual tokens to behave more like natural language, which leads to enhanced performance in both visual comprehension and generation. 

\begin{table}[t]
    \centering
    \begin{adjustbox}{width=0.5\textwidth}
    \begin{tabular}{ll|ccccccc}
    \toprule
   & Models &   MMD & VST & VRI & MMC & KGQA & TRQA & MMR  \\ \hline
      \multicolumn{2}{l|}{Morph-token}   &  \textbf{32.2}   &   \textbf{27.4}   &   \textbf{27.4}   &  \textbf{28.0}  &  \textbf{56.4}  &   \textbf{47.7}  &  \textbf{54.9} \\ 
    1& detail-detail   &   7.3   &   8.6   &   8.2   &   16.6  &  30.2  &   24.8  &  34.8\\ 
    2& abstr-abstr (SD)   &    25.8  &   17.3   &   11.3   &   22.7  &  38.8  & 33.5   &  45.0 \\ 
     3& abstr-abstr (VQ)   &  25.2    &  20.4  & 13.2   &   19.8 & 37.9    &  32.7 &  46.1 \\ \hline
    4& w/o decoder   &   20.9   &   16.4  &   12.8   &   18.0 &  35.6  &   30.4  &  41.3 \\ 
    5& w/o deconfound    &   31.8   &   25.7   &   25.3   &   26.2  &  54.7  &   44.8  &  50.9 \\ \hline
    6& continuous &   31.7   &   26.3   &   26.1   &  27.0  &  54.3  &   45.2  &  52.5 \\
    \bottomrule
    \end{tabular}
    \end{adjustbox}
    \vspace{-1em}
    \caption{\label{tab:abla-demon} Ablation results on DEMON Benchmark.}
\end{table}

\begin{table}[t]
    \centering
    \begin{adjustbox}{width=0.38\textwidth}
    \begin{tabular}{ll|ccc}
    \toprule
  &Models &   \textbf{EVR}$\downarrow$ & \textbf{MA5K}$\downarrow$ & \textbf{MagicBrush}$\downarrow$ \\ \hline

  \multicolumn{2}{l|}{Morph-token}        & \textbf{15.3}      & \textbf{14.6}     & \textbf{7.6}     \\
  1& detail-detail  & 34.1     & 31.2         & 32.5 \\
    2& abstr-abstr (SD)  & 27.7     & 26.6      & 27.8 \\
    3& abstr-abstr (VQ)   & 25.2     & 24.4      & 23.9 \\ \hline
    4& w/o decoder   & 31.7     & 28.9      & 30.3 \\
    5& w/o deconfound   & 18.5     & 18.2      & 12.2 \\ \hline
    6& continuous   & 16.1     & 15.8       & 9.7 \\
   
    \bottomrule
    \end{tabular}
    \end{adjustbox}
    \vspace{-0.5em}
    \caption{\label{tab:abla-edit}Ablation results on image editting with L1 score as the evalution metric.}
    \vspace{-1em}
\end{table}
\paragraph{Imapct of Discrete Morph-Tokens.}
We also validate the superiority of quantizing visual abstraction into discrete tokens as the pre-MLLM morph-tokens.
We eliminate the process of visual quantization and also change the optimization objective of visual tokens as regressing the next token with MSE loss in stage 1.
As indicated in Row 6 of Table~\ref{tab:abla-demon} and \ref{tab:abla-edit}, it results in a degradation of model performance for both comprehension and generation tasks. 
We argue a key reason for this is that discretization aligns visual tokens more closely with the attributes of natural language.
Then in training stage 1, we can employ a uniform objective (cross-entropy loss) for both vision and language, which better improves their alignment within the LLM~\cite{jin2023unified}, thereby facilitating a more effective transition of the LLM into an MLLM.

\begin{figure}[t]
    \centering
    \includegraphics[width=\linewidth]{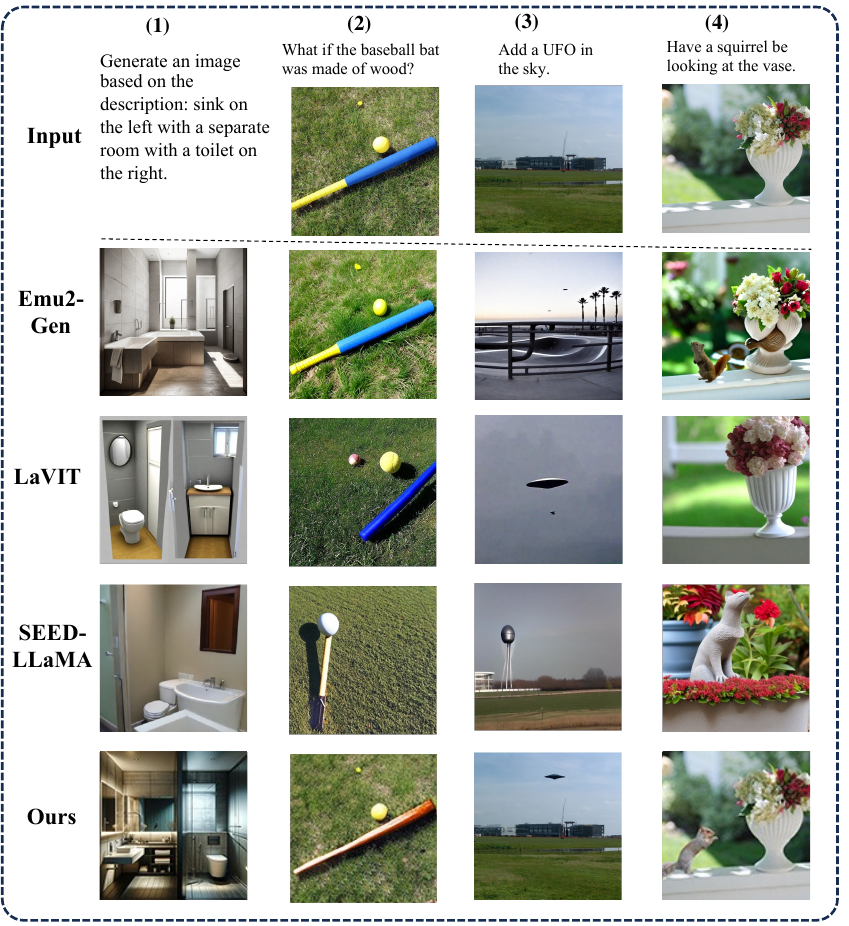}
    \vspace{-1.4em}
    \caption{ Qualitative comparison on image synthesis.}
    \label{fig:gen_comp}
    \vspace{-0.5em}
\end{figure}

\paragraph{Image-Text Retrieval with our Encoder.}
To further validate that our proposed encoder can effectively translate the non-linguistic image into a sequence of morph-tokens  that behave more like natural language, 
we further evaluate the performance of our encoder in zero-shot image-text retrieval tasks, utilizing the Flickr30K dataset with Recall@K (R@K) as the evaluation metric.
We compare with the visual encoders in SEED-LLaMA~\cite{ge2023planting} and LAVIT~\cite{jin2023unified}, as well as the Qformer in BLIP2~\cite{li2023blip}.
As shown in Table~\ref{tab:retrieval}, our encoder surpasses those from existing MLLMs across the majority of metrics in text-image retrieval tasks, while also achieving superior average performance (R@m). 
This validates that our encoder can better abstract visual semantics to align with text representations, thereby effectively alleviating the modality gap.

\paragraph{Qualitative Comparison on Image Synthesis.}
Figure~\ref{fig:gen_comp}  presents a qualitative comparison between our model and other MLLMs in the context of image generation tasks.
We can see that in the text-to-image generation task when the textual descriptions involve spatial relationships, images generated by MLLM often inaccurately represent the relationships among specific visual objects. In contrast, our model precisely generates the image that adheres to the predetermined specifications detailed in the prompt, reflecting the intended spatial relationships.
While in image editing scenarios, it can be observed that our approach well understands the instructions, while also effectively preserving image fidelity, which is rarely observed in prior works.

\section{Conclusion}

In this paper, we propose Morph-Tokens to resolve the conflicting training objectives between visual comprehension and generation-- ``morph''  implies a transformation where the pre-MLLM visual tokens are not necessarily equal to the post-MLLM ones.  The pre-MLLM tokens are abstract semantics, serving as visual prompts for comprehension tasks while the post-MLLM tokens
are visually complete tokens for image generation. We further propose a 3-stage training strategy, detaching the textual and image reconstruction losses with our morph-tokens. After training, our model showcases notable zero-shot performance on a broad range of comprehension and generation tasks, also exhibiting extensive emergent abilities such as consistently preserving image
fidelity in image editing scenarios.

\section{Impact Statements}
\textbf{Ethical Impacts} This study does not raise any ethical concerns. The research does not involve subjective assessments or the use of private data. Only publicly available datasets are utilized for experimentation.

\textbf{Expected Societal Implications}.
A major societal concern with this technology lies in its potential for misuse, particularly in fabricating unauthorized images that could lead to misinformation, privacy breaches, and other damaging consequences. To counter these threats, it is crucial to develop strong ethical standards and implement ongoing surveillance.

The issue highlighted is not unique to our method but is prevalent across different techniques for multi-concept customization. A practical approach to mitigating these risks could involve the use of adversarial technologies designed to reverse unauthorized modifications. Additionally, embedding imperceptible watermarks in created images could act as a preventative measure against abuse and guarantee that their use is properly attributed.

\nocite{langley00}

\bibliography{example_paper}
\bibliographystyle{icml2024}

\newpage
\appendix
\onecolumn
\section{Detailed Comparison with Existing MLLMs.}
\label{sec:A}
MLLMs~\cite{li2023blip, li2022fine, zhu2023minigpt} aim to serve as a general-purpose assistant to perform various vision-language tasks with strong generalization ability~\cite{zhang2024out, pan2023self}.
To drive both multimodal comprehension and generation tasks in a unified, token-based, auto-regressive framework, existing approaches~\cite{sun2023generative,yang2023teal,ge2023planting, ge2024worldgpt, jin2023unified} typically leverage a tokenizer to encode images into visual tokens and feed them into an MLLM for vision-language comprehension. 
In terms of visual generation, the post-MLLM visual tokens are further used to generate target images, which is achieved either through a pre-trained VQVAE decoder dedicated to image pixel reconstruction, or via Stable Diffusion (SD~\cite{rombach2021highresolution}) models. 

Specifically, models like TEAL~\citep{yang2023teal} and VideoPoet~\citep{kondratyuk2023videopoet}, utilize the VQ-VAE encoder as the tokenizer (e.g., VQGAN~\cite{esser2021taming} and MAGVIT-v2\cite{yu2023language}), encoding images into visual tokens with detailed semantics for unified auto-regression within MLLMs. 
The post-MLLM visual tokens, which are visually complete, can then be directly converted into images using the corresponding VQ-VAE decoder. 
However, as these tokens preserve low-level visual details, while being suitable for visual generation, they substantially impede the capability of visual comprehension.

Another line of work~\citep{ge2023planting, sun2023generative} attempts to first extract abstracted visuals for comprehension, where visual tokens and text tokens undergo a unified autoregression within the MLLM.  
And then akin to ``Textual Inversion''~\cite{gal2022image}, the post-MLLM visual tokens are further aligned into the condition embedding space of existing SD model (e.g., through MSE loss), facilitating SD models to generate the image.
However, two significant challenges arise: (1) post-MLLM visual tokens, similar to pre-MLLM ones, also encapsulate abstract semantics that are insufficient for image generation. 
To address the conflicting objectives, methods like Emu2~\cite{Emu2} opt to train separate models for distinct purposes: Emu-gen for generation and Emu-chat for comprehension.
(2) Moreover, existing SD models mainly focus on simple scene image generation, trained with coarse-grained conditions~\cite{li2022upainting}. For instance, the unCLIP-SD~\cite{rombach2022high} used in SEED-LLaMA~\cite{ge2023planting} utilizes CLIP image embeddings as the condition, which contain only modality-shared information and often overlook the modality-specific knowledge derived from multimodal comprehension~\citep{liang2022mind, dong2023dreamllm}. 
Therefore, it is challenging for SD models to achieve detailed control during image generation which rarely preserves image fidelity, especially in image editing scenarios.
And comparing the results of rows 2 and 3 in Table~\ref{tab:abla-edit} further substantiates this point. 
We can see that introducing an additional decoder for image reconstruction (Row3 in Table~\ref{tab:abla-edit}) yields better image editing performance (measured by L1 score) compared to simply aligning with the modality-shared condition embedding of SD models (Row2 in Table~\ref{tab:abla-edit}), where the decoder allows the lower-level visual tokens to autoregressively generate their own visual distributions. 

In contrast to these MLLMs, we propose morph-tokens to detach the textual and image reconstruction losses, where the pre-MLLM visual tokens (with abstract semantics) are not necessarily equal to the post-MLLM ones (with visually-complete semantics), effectively resolving the conflicting objectives between comprehension and generation.
Moreover, we employ a deconfounded Qformer as the encoder, enabling pre-MLLM visual tokens to behave more like natural language compared with existing tokenizers (e.g., SEED\citep{ge2023planting}).
And we further introduce another decoder to alleviate the burden of MLLM for visual semantic recovering, consequently fostering a synergy between visual comprehension and generation.

\section{Detailed Implementations of our Framework}
\label{sec:B}
We mainly introduce the detailed implementations of the encoder.
Given an image, it is first transformed into a sequence of visual tokens $\mathcal{V}$ via CLIP-ViT, with each token encapsulating patch-level visual details.
And the role of the encoder is to abstract these visuals by transforming them into morph-tokens.
To achieve this, we propose a novel deconfounded Qformer to implement our encoder, eliminating the spatially spurious correlation in vision, enabling the resultant morph-tokens to behave more like natural language.

Firstly, following Qformer, we introduce a set of learnable query vectors (here we refer to as group tokens), and employ an attention-based method for semantic aggregation. 
Specifically, in contrast to the vanilla Qformer, we implement two improvements:
we upgrade the self-attention mechanism to causal self-attention, wherein each token exclusively attends to its preceding tokens, thus endowing the sequence with  causal dependency. 
Furthermore, we replace the pivotal cross-attention computation in Qformer with slot-attention, which still utilizes the group tokens as the query and visual tokens as the key/value. Diverging from the traditional cross attention in transformer decoders, slot-attention performs normalisation over queries, encouraging each visual token to be claimed by one of the group tokens, with the attention score $\mathbb{A}$ calculated as follows:
\begin{equation}
\label{eq:aggregation3}
\small
\begin{aligned}
\mathbb{A} = \text{Softmax}_{qry}(f_{\mathcal{G}}(X)) 
= \text{Softmax}_{qry}(\frac{(\mathcal{G}W_q)(XW_k)^T}{\sqrt{scaled}}),\ \  \sum_j{\mathbb{A}{j,k}}=1
\end{aligned}
\end{equation}
And then the output of the Qformer successfully encapsulates the desired visual abstraction. Post-processing through an additional MLP layer, the features of visual abstraction are then passed to a learnable codebook $\mathcal{C}$ and quantized into a sequence of discrete visual codes   as the morph-tokens through nearest neighbors lookup.

Based on the above framework, we further introduce the design of deconfounding to enhance morph-tokens for emulating natural language.
Unlike the sequential manner in which humans understand language, image comprehension typically involves capturing a holistic visual impression from several key areas, and then diverging into specific image details.
Sequentially flattening 2D images into 1D features can result in spurious correlations between two spatial visual tokens, thereby confounding the semantic abstraction of specific visual objects.
For instance, imagine an image where a boy is leisurely watching a disaster movie at home. Reading the image sequentially akin to text processing may lead to a confused understanding of virtual and real worlds, mistakenly placing the boy within the movie scene, consequently extracting incorrect information about him.

\begin{wrapfigure}{r}{0.4\textwidth}
\vspace{-0.6cm}
\begin{center}
    \includegraphics[width=0.4\textwidth]{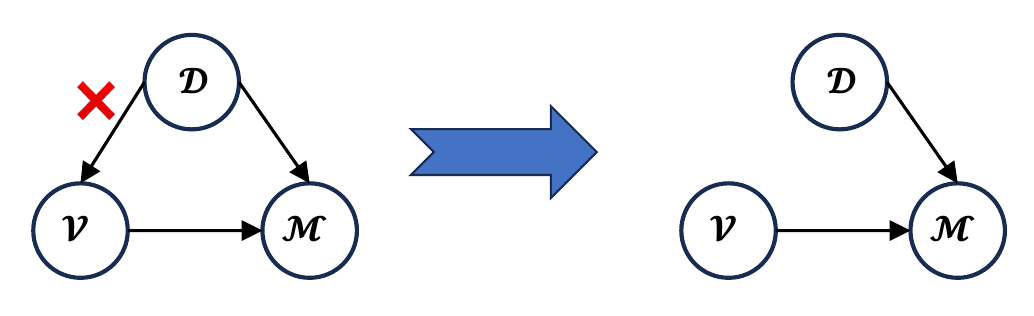}
    \caption{causal intervention.}
    \label{causal}
\end{center}
\end{wrapfigure} 

In order to screen out the existence of confounders and then eliminate their effect, we use a mental apparatus, \textbf{intervention}. 
As depicted in Figure \ref{causal}, $\mathcal{V}$ corresponds to the  visual tokens with detailed patch-level semantics, and $\mathcal{M}$ represents the abstracted visual token after semantic aggregation. $\mathcal{V}\rightarrow \mathcal{M}$ denotes the process of visual semantic abstraction. 
Additionally, the confounder $\mathcal{D}$ represents other image patches which also directly affect $\mathcal{M}$ during aggregation. Simultaneously, its existence may erroneously impact the semantic abstraction for $\mathcal{V}$, leading to spurious correlations by relying solely on the likelihood  $P(\mathcal{M}|\mathcal{V})$, where the confounder introduces the observational bias via $P(\mathit{d}|\mathcal{V})$:
\begin{equation}
\small
\label{eq:causal1}
\begin{aligned}
    P(\mathcal{M}|\mathcal{V}) := \sum_{\mathit{d} \in \mathcal{D}} P(\mathcal{M}|\mathcal{V},\mathit{d})\underline{P(\mathit{d} | \mathcal{V})}
\end{aligned}
\end{equation}
To calculate the true causal effect between $\mathcal{V}$ and $\mathcal{M}$, we could intervene on $\mathcal{V}$ to cut off the causal link between $\mathcal{D}$ and $\mathcal{V}$, as shown in Figure~\ref{causal}(right). Utilizing Bayes rule on the revised graph, we have:
\begin{equation}
\small
\label{eq:causal2}
\begin{aligned}
    P(\mathcal{M}|do(\mathcal{V})) := \sum_{\mathit{d} \in \mathcal{D}} P(\mathcal{M}|\mathcal{V},\mathit{d})\underline{P(\mathit{d})}
\end{aligned}
\end{equation}
In contrast to Eq.~(\ref{eq:causal1}), $\mathit{d}$ is no longer influenced by $\mathcal{V}$. Consequently, the intervention compels $\mathcal{V}$ to fairly incorporate every $\mathit{d}$ into the prediction of $\mathcal{M}$, in accordance with the prior probability $P(\mathit{d})$.

To implement the causal intervention in Eq.~(\ref{eq:causal2}), we should first include an additional confounder set $\mathcal{D}$ to enumerate the feature of different image patches, while it is impossible to attend to all image pathes in the real world. Fortunately, some works, \textit{e.g.}, VIT-VQGAN~\cite{yu2021vector} or DALLE~\cite{ramesh2021zero}, effectively quantize path-level embeddings from diverse images into a finite set of discretized latent codes within a learned codebook, providing a valuable resource to initialize the confounder dictionary.

As shown in Eq.~(\ref{eq:aggregation3}), the critical step of visual semantic abstraction ($\mathcal{V}\rightarrow \mathcal{M}$) is facilitated through the slot attention, with a query-wise softmax is used to determine which clusters a low-level detailed visual token should be allocated to. Therefore, the implementation of causal intervention should be reflected in slot attention to upgrade the query-wise softmax, as delineated below:
\begin{equation}
\small
\label{eq:causal3}
\begin{aligned}
   P(\mathcal{M}|do(\mathcal{V})) := \mathbb{E}_\mathbf{d}[\text{Softmax}_{qry}(f_{\mathcal{G}}(\mathcal{V}, \mathit{d}))] 
\end{aligned}
\end{equation}
where $f(\cdot)$ calculates the logits of scaled dot-product attention. However, in this way $\mathbb{E}_\mathbf{d}$ requires expensive sampling with the cost of a network forward pass for all terms in $\mathcal{D}$. So we apply Normalized Weighted Geometric Mean (NWGM) to approximate the above expectation, effeciently moving the outer expectation into the Softmax operation as:
\begin{equation}
\small
\label{eq:causal4}
\begin{aligned}
    \mathbb{E}_\mathbf{z}[\text{Softmax}_{qry}(f_{\mathcal{G}}(\mathcal{V},d))] \approx \text{Softmax}_{qry}(\mathbb{E}_\mathbf{d}[f_{\mathcal{G}}(\mathcal{V},d)])
\end{aligned}
\end{equation}
To achieve the above approximation, we expand the query of slot attention as $\mathcal{Q} = \mathcal{G}W_q + \mathbb{E}_d[h_{\mathcal{G}}(d)]$, and then Eq.~(\ref{eq:causal4}) can be derived as:
\begin{equation}
\small
\begin{aligned}
    &\mathbb{E}_\mathbf{z}[\text{Softmax}_{qry}(f_{\mathcal{G}}(\mathcal{V},d))] \\
     &= \text{Softmax}_{qry}(\frac{(\mathcal{G}W_q + \mathbb{E}_\mathbf{d}[h_{\mathcal{G}}(d)])(\mathcal{V}W_k)^T}{\sqrt{scaled}})
\end{aligned}
\end{equation}
Moreover, to compute $\mathbb{E}_\mathbf{d}[h_{\mathcal{G}}(d)]$, we also employ an attention-based mechanism, which, for convenience, we refer to as a single-layer Q-former (a module that includes only the computation of cross-attention).
We treat the group tokens $\mathcal{G}$ as the query and the confounder dictionary $\mathcal{D}$ as both key and value. 
Through this, we derive an attention matrix $\mathcal{A}$ over each item in the dictionary. Then we can have $\mathbb{E}_\mathbf{d}[h_{\mathcal{G}}(d)] = \sum_z [A \odot \mathcal{D}]P(d)$, where $P(d)$ signifies the prior statistical probability and $\odot$ represents the element-wise product.

\section{Experimental Details}
\label{sec:C}
\subsection{Data}
\paragraph{Pretraining Data. }
In stage 1 and stage 2, we select $\sim$30M image-text pairs from CC3M~\cite{sharma2018conceptual} and Laion~\cite{laion-coco}, which are concatenated in two formats, \textit{i.e.}, [text][image] and [image][text], facilitating the alignment between text and vision.

\paragraph{Instruction Tuning Data. } 

During instruction tuning~\cite{liu2023visual,zhang2024recost}, we incorporate a variety of tasks, outlined as follows:
(1) Text-to-Image Generation: We employ datasets including JourneyDB~\cite{pan2023journeydb} and DiffusionDB~\cite{wang2022diffusiondb}, utilizing a prompt template formatted as:``\texttt{USER: \{caption\} Generate an image based on the description. ASSISTANT: \{image\}}''.

(2) Image editing:
We employ datasets such as IPr2Pr~\cite{brooks2023instructpix2pix},  utilizing a prompt template formatted as:``\texttt{USER: \{image1\} What will this image be like with the editing instruction: \{instruction\}. ASSISTANT: \{image2\}}''.

(3) Image caption \& Image QA \& Video QA: We mainly leverage the held-in instruction-tuning datasets and corresponding instruction templates used in InstructBlip~\cite{instructblip}.

(4) Image Conversation: We employ the datasets including LLaVA~\cite{liu2023visual}, SVIT~\cite{zhao2023svit} with the prompt template formatted as: ``\texttt{USER: \{image\} \{question\}. ASSISTANT: \{answer\}}''.

(5) Multi-Image Understanding: We leverage GSD~\cite{li2023mimic} as the training dataset, utilizing a prompt template formatted as:``\texttt{USER: This is the first image. \{image1\} This is the second image. \{image2\} \{question
\} ASSISTANT: \{answer\}}''. 

\paragraph{Evaluation Data. }
For comprehension tasks,  we first evaluate on a wide range of academic benchmarks, including NoCaps~\cite{nocaps}, Flickr30K~\cite{young-etal-2014-image}, GQA~\cite{gqa}, VSR~\cite{vsr}, ICONQA~\cite{iconqa}, HatefulMeme~\cite{hatefulmemes}, MSVDQA~\cite{msvdqa}, and MSRVTTQA~\cite{msvdqa}.
The split of test sets and the evaluation metrics are aligned with those described in InstructBlip~\cite{instructblip}. 
Additionally, we also include some MLLM-oriented comprehension benchmarks, such as MME~\cite{fu2023mme} and the DEMON benchmark~\cite{li2023finetuning}.
For generation tasks, our evaluation encompasses both text-to-image generation and image editing. The former includes datasets of MS-COCO~\cite{lin2014microsoft}, (with 30K randomly sampled data from the validation set and 5K data from the Karpathy test set), and Flickr30K~\cite{young-etal-2014-image} (with 1K data in the test set).
For image editing, we evaluate the performance using datasets such as EVR~\cite{tan2019expressing}, MA5k\cite{shi2021learning}, and MagicBrush~\cite{zhang2023magicbrush}. The partitioning of test sets and the evaluation metrics adhere to MGIE~\cite{fu2023guiding}.

\subsection{Training.} We train the entire set of parameters for both the encoder and decoder. For the LLM, to enhance efficiency, we employ 
LoRA tuning~\cite{hu2021lora} and together optimize the parameters of the decoder head layer due to the added visual words. 
With LoRA, we finetune $W_q$ and $W_v$ via low-rank adaptation. In our
implementation, we set the rank, $r = 64$.
utilize the AdamW optimizer coupled with a cosine learning rate scheduler. The hyperparameters for the AdamW optimizer are set with 
$\beta=(0.9, 0.999)$, and we apply a weight decay of 0.05.
The training is conducted on 16xA800 GPUs.
For the first two stages, we train for 200,000 steps with a maximum learning rate as 1e-4. During instruction tuning, the model is trained for 100,000 steps with a maximum learning rate of 1e-5.

\subsection{Ablation Model Implementation.}
Here we give some implementation details of ablation models.
(1) detail-detail: both pre- and post-MLLM visual tokens contain detailed semantics. 
Following TEAL, We integrate visual tokens from VQ-GAN encoder into the MLLM for unified auto-regression alongside text tokens, and directly convert the post-MLLM tokens into a specific image via VQ-GAN decoder.
(2) abstr-abstr (SD): Both pre- and post-MLLM visual tokens contain abstract semantics. 
Utilizing our encoder to abstract the visuals, we leverage a unified auto-regressive objective for both textual and visual tokens within MLLM. Furthermore, following SEED-LLaMA~\cite{ge2023planting}, the post-MLLM visual tokens are aligned with the condition embedding of SD model~\cite{rombach2022high} (trained with MSE loss between token-embedding and the ground-truth condition embedding) 
(3) abstr-abstr (VQ):  Based on the previous ablation model, we remove the SD model and instead train a decoder-only transformer, which auto-regressively predicts the complete visual token sequence that can be decoded into an image via VQGAN decoder~\cite{esser2021taming}, instructed by post-MLLM visual tokens.

Moreover, we also conduct image-text retrieval experiments~\cite{cao2022imagetext, pan2024i3}. We adopt the dual-stream paradigm and incorporate the text encoder from BLIP2~\cite{li2023blip}. 
Concurrently, we learn a linear-projection layer using some image-text pairs from LAION, to align the output of the morph-encoder with that of the text encoder.
We compare with the visual encoders in SEED-LLaMA~\cite{ge2023planting} and LAVIT~\cite{jin2023unified}, as well as the Qformer in BLIP2 (we remove the image-text-matching re-rank module in BLIP2 to ensure a fair comparison).


\end{document}